\documentclass[]{llncs}
\usepackage{graphicx}
\usepackage{amsmath,amssymb} 
\usepackage{color}
\usepackage[width=122mm,left=12mm,paperwidth=146mm,height=193mm,top=12mm,paperheight=217mm]{geometry}

\usepackage{epstopdf}
\usepackage{booktabs}
\usepackage{adjustbox}
\usepackage[pagebackref=true,breaklinks=true,colorlinks,bookmarks=false,urlcolor=blue]{hyperref}
\usepackage{cite} 
\usepackage{subfig}
\usepackage{xspace}

\DeclareMathOperator*{\argmax}{arg\,max}

\makeatletter
\DeclareRobustCommand\onedot{\futurelet\@let@token\@onedot}
\def\@onedot{\ifx\@let@token.\else.\null\fi\xspace}
\def\eg{\emph{e.g}\onedot} 
\def\ie{\emph{i.e}\onedot} 
\def\etal{\emph{et al}\onedot}
\makeatother

\newcommand{\startcompact}[1]{\par\vspace{-0.75em}\begin{#1}%
\allowdisplaybreaks\ignorespaces}
\newcommand{\stopcompact}[1]{\end{#1}\ignorespaces}

\begin{document}
\pagestyle{headings}

\title{Learning Free-Form Deformations \\ for 3D Object Reconstruction} 



\author{Dominic Jack, Jhony K. Pontes, Sridha Sridharan, Clinton Fookes, Sareh Shirazi, Frederic Maire, Anders Eriksson}
\institute{Queensland University of Technology, Brisbane, Australia}


\maketitle


\begin{abstract}
Representing 3D shape in deep learning frameworks in an accurate, efficient and compact manner still remains an open challenge. Most existing work addresses this issue by employing voxel-based representations. While these approaches benefit greatly from advances in computer vision by generalizing 2D convolutions to the 3D setting, they also have several considerable drawbacks. The computational complexity of voxel-encodings grows cubically with the resolution thus limiting such representations to low-resolution 3D reconstruction. In an attempt to solve this problem, point cloud representations have been proposed. Although point clouds are more efficient than voxel representations as they only cover surfaces rather than volumes, they do not encode detailed geometric information about relationships between points. In this paper we propose a method to learn free-form deformations (\textsc{Ffd}) for the task of 3D reconstruction from a single image. By learning to deform points sampled from a high-quality mesh, our trained model can be used to produce arbitrarily dense point clouds or meshes with fine-grained geometry. We evaluate our proposed framework on both synthetic and real-world data and achieve state-of-the-art results on point-cloud and volumetric metrics. Additionally, we qualitatively demonstrate its applicability to label transferring for 3D semantic segmentation.\footnote{Code available at \href{https://github.com/jackd/template_ffd}{github.com/jackd/template\_ffd}}
\end{abstract}

\section{Introduction}
Imagine one wants to interact with objects from the real world, say a chair, but in an augmented reality (AR) environment. The 3D reconstruction from the seen images should appear as realistic as possible so that one may not even perceive the chair as being virtual. The future of highly immersive AR and virtual reality (VR) applications highly depends on the representation and reconstruction of high-quality 3D models. This is obviously challenging and the computer vision and graphics communities have been working hard on such problems \cite{Penner2017soft3d,huang2015,maier2017intrinsic3d}.

The impact that recent developments in deep learning approaches have had on computer vision has been immense. In the 2D domain, convolutional neural networks (CNNs) have achieved state-of-the-art results in a wide range of applications \cite{NIPS2012_4824,farabet2013learning,graves2009novel}. Motivated by this, researchers have been applying the same techniques to represent and reconstruct 3D data. Most of them rely on volumetric shape representation so one can perform 3D convolutions on the structured voxel grid \cite{choy2016,Yan2016,QiSu2016,Kar2017,Zhu2017,marrnet2017}.
A drawback is that convolutions on the 3D space are computationally expensive and grow cubically with resolution, thus typically limiting the 3D reconstruction to exceedingly coarse representations.

Point cloud representation has recently been investigated to make the learning more efficient \cite{fan2016point,Qi_CVPR2017,Qi2017,lin2017learning}. However, such representations still lack the ability of describing finely detailed structures. Applying surfaces, texture and lighting to unstructured point clouds are also challenging, specially in the case of noisy, incomplete and sparse data.

The most extensively used shape representation in computer graphics is that of polygon meshes,
in particular using triangular faces. This parameterisation has largely been unexplored in the machine
learning domain for the 3D reconstruction task. This is in part a consequence of most
machine learning algorithms requiring regular representation of input and output data such as voxels and point clouds. Meshes are highly unstructured and their topological structure usually differs from one to another which makes their 3D reconstruction from 2D images using neural networks challenging.

\begin{figure}[ht]
\setlength{\hsize}{\textwidth}
\centering
\includegraphics[width=0.7\textwidth,keepaspectratio]{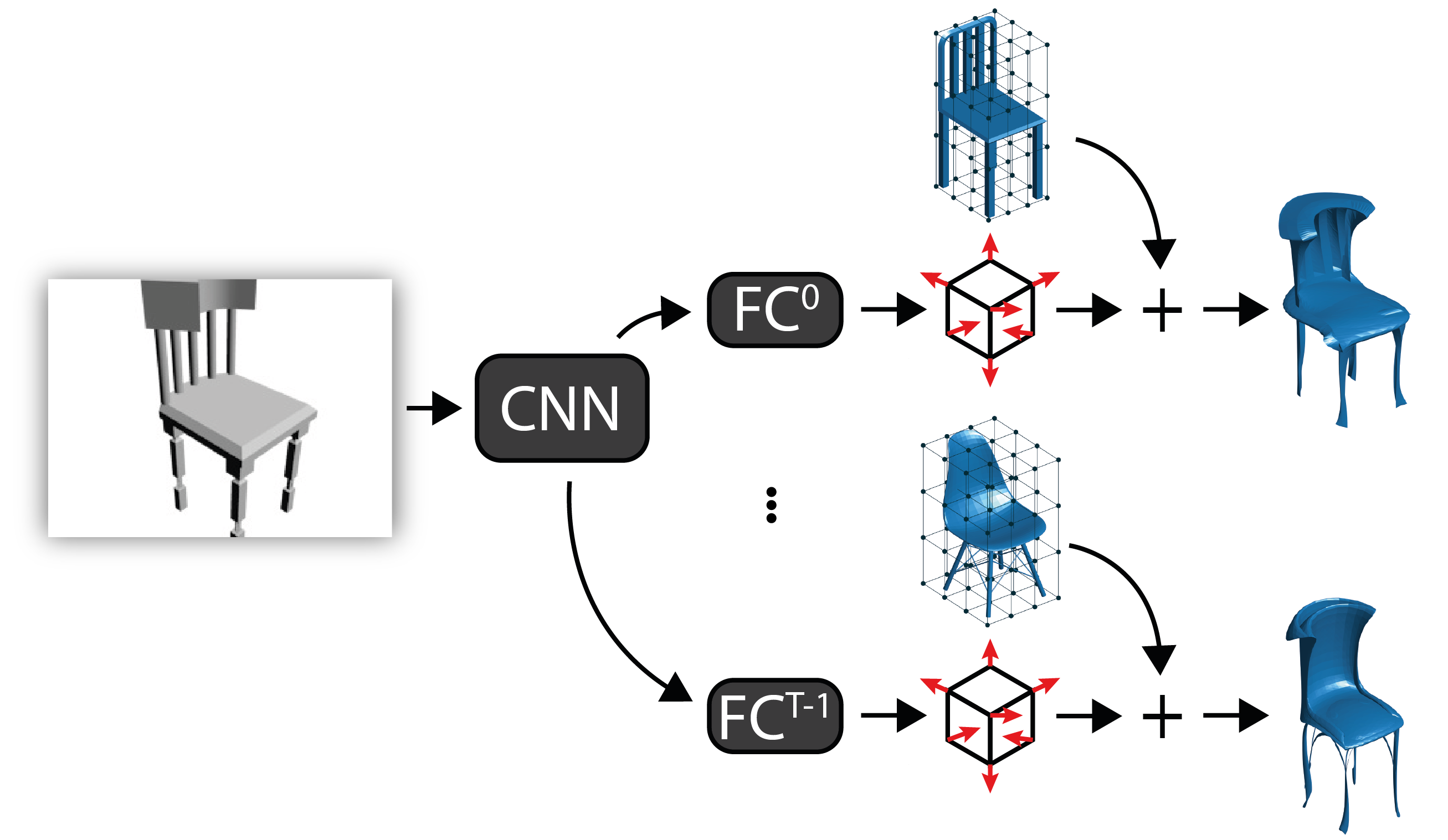}
\caption{Given a single image, our method uses a CNN to infer \textsc{Ffd} parameters $\Delta \mathbf{P}$ (red arrows) for multiple templates $T$ (middle meshes). The $\Delta \mathbf{P}$ parameters are then used to deform the template vertices to infer a 3D mesh for each template (right meshes). Trained only with surface-sampled point-clouds, the model learns to apply large deformations to topologically different templates to produce inferred meshes with similar surfaces. Likelihood weightings $\gamma$ are also inferred by the network but not shown for simplicity. FC stands for fully connected layer.}
\label{fig:overview}
\end{figure}

In this paper, we tackle this problem by exploring the well-known free-form deformation (\textsc{Ffd}) technique \cite{Sederberg1986} widely used for 3D mesh modelling. \textsc{Ffd} allows us to deform any 3D mesh by repositioning a few predefined control points while keeping its topological aspects. We propose an approach to perform 3D mesh reconstruction from single images by simultaneously learning to select and deform template meshes. Our method uses a lightweight CNN to infer the low-dimensional \textsc{Ffd} parameters for multiple templates and it learns to apply large deformations to topologically different templates to produce inferred meshes with similar surfaces. We extensively demonstrate relatively small CNNs can learn these deformations well, and achieve compelling mesh reconstructions with finer geometry than standard voxel and point cloud based methods. An overview of the proposed method is illustrated in Figure~\ref{fig:overview}. Furthermore, we visually demonstrate our proposed learning framework is able to transfer semantic labels from a 3D mesh onto unseen objects.

\noindent\textbf{\\Our contributions are summarized as follows:}
\begin{itemize}
\item We propose a novel learning framework to reconstruct 3D meshes from single images with finer geometry than voxel and point cloud based methods;
\item we quantitatively and qualitatively demonstrate that relatively small neural networks require minimal adaptation to learn to simultaneously select appropriate models from a number of templates and deform these templates to perform 3D mesh reconstruction;
\item we extensively investigate simple changes to training and loss functions to promote variation in template selection; and
\item we visually demonstrate our proposed method is able to transfer semantic labels onto the inferred 3D objects.
\end{itemize}

\section{Related Work}
Interest in analysing 3D models has increased tremendously in recent years. This development has been driven in part by a rapid growth of the amount of readily available 3D data, the astounding progress made in the field of machine learning as well as a substantial rise in the number of potential
application areas, \ie Virtual and Augmented Reality.

To address 3D vision problems with deep learning techniques a good shape representation should be found. Volumetric representation has been the most widely used for 3D learning \cite{Ulusoy2015,Wu2015,Cherabier2016,choy2016,Sharma2016,Rezende2016,Yan2016,QiSu2016,marrnet2017,Kar2017,Zhu2017}. Convolutions, pooling, and other techniques that have been successfully applied to the 2D domain can be naturally applied to the 3D case for the learning process. Volumetric autoencoders \cite{Girdhar2016,Sharma2016} and generative adversarial networks (GANs) have been introduced \cite{Wu2016,Liu2017,Gwak2017} to learn probabilistic latent space of 3D objects for object completion, classification and 3D reconstruction. Volumetric representation however grows cubically in terms of memory and computational complexity as the voxel grid resolution increases, thus limiting it to low-quality 3D reconstructions.

To overcome these limitations, octree-based neural networks have been presented \cite{Riegler2017,Wang2017,Hane2017,Maxim2017}, where the volumetric grid is split  recursively by dividing it into octants. Octrees reduce the computational complexity of the 3D convolution since the computations are focused only on regions where most of the object's geometry information is located. They allow for higher resolution 3D reconstructions and a more efficient training, however, the outputs still lack of fine-scaled geometry. A more efficient 3D representation using point clouds was recently proposed to address some of these drawbacks \cite{fan2016point,Qi_CVPR2017,Qi2017,lin2017learning}. In \cite{fan2016point}
a generative neural network was presented to directly output a set of unordered 3D points that can be used for the 3D reconstruction from single image and shape completion tasks. By now, such architectures have been demonstrated for the generation of relatively low-resolution outputs and to scale these networks to higher resolution is yet to be explored.

3D shapes can be efficiently represented by polygon meshes which encode both geometrical (point cloud) and topological (surface connectivity) information. However, it is difficult to parametrize meshes to be used within learning frameworks \cite{Lun3DV2017}. A deep residual network to generate 3D meshes has been proposed in \cite{Sinha2017}. A limitation however is that they adopted the geometry image representation for generative modelling of 3D surfaces so it can only manage simple (\ie genus-0) and low-quality surfaces. In \cite{huang2015}, the authors reconstruct 3D mesh objects from single images by jointly analysing a collection of images of different objects along with a smaller collection of existing 3D models. While the method yields impressive results, it suffers from scalability issues and is sensitive to semantic segmentation of the image and dense correspondences.

\textsc{Ffd} has also been explored for 3D mesh representation where one can intrinsically represent an object by a set of polynomial basis functions and a small number of coefficients known as control points used for cage-like deformation. A 3D mesh editing tool proposed in \cite{Yumer2016} uses a volumetric generative network to infer per-voxel deformation flows using \textsc{Ffd}. Their method takes a volumetric representation of a 3D mesh as input and a high-level deformation intention label (\eg sporty car, fighter jet, etc.) to learn the \textsc{Ffd} displacements to be applied to the original mesh. In \cite{Chen2017,Pontes2017} a method for 3D mesh reconstruction from a single image was proposed based on a low-dimensional parametrization using \textsc{Ffd} and sparse linear combinations given the image silhouette and class-specific landmarks. Recently, the DeformNet was proposed in \cite{kuryenkov2017deformnet} where they employed \textsc{Ffd} as a differentiable layer in their 3D reconstruction framework. The method builds upon two networks, one 2D CNN for 3D shape retrieval and one 3D CNN to infer \textsc{Ffd} parameters to deform the 3D point cloud of the shape retrieved. In contrast, our proposed method reconstructs 3D \textit{meshes} using a single lightweight CNN with \textit{no} 3D convolutions involved to infer a 3D mesh template and its deformation flow in one shot.

\section{Problem Statement}
We focus on the problem of inferring a 3D mesh from a single image.
We represent a 3D mesh $c$ by a list of vertex coordinates
$\mathbf{V} \in \mathbb{R}^{n_v \times 3}$ and a set of triangular faces $\mathbf{F} \in \mathbb{Z}^{n_f \times 3}, \, 0 \le f_{ij} < n_v$ defined such that $\mathbf{f}_i = [p, q, r]$ indicates there is a face connecting the vertices $\mathbf{v}_p$, $\mathbf{v}_q$ and $\mathbf{v}_r$, \ie $c = \{\mathbf{V}, \mathbf{F}\}$.

Given a query image, the task is to infer a 3D mesh $\tilde{c}$ which is close by some measure to the actual mesh $c$ of the object in the image. We employ the \textsc{Ffd} technique to deform the 3D mesh to best fit the image.

\subsection{Comparing 3D Meshes}
There are a number of metrics which can be used to compare 3D meshes. We consider three: Chamfer distance and earth mover distance between point clouds, and the intersection-over-union (IoU) of their voxelized representations.

\paragraph{\textbf{Chamfer distance.}} The Chamfer distance $\lambda_c$  between two point clouds $A$ and $B$ is defined as

\startcompact{small}
\begin{equation}
\lambda_c (A, B) = \sum_{a \in A} \min_{b \in B}  \Vert a - b \Vert^2 + \sum_{b \in B} \min_{a \in A} \Vert b - a \Vert^2.
\label{eqn:chamfer2}
\end{equation}
\stopcompact{small}

\paragraph{\textbf{Earth mover distance.}} The earth mover \cite{rubner2000earth} distance $\lambda_{em}$ between two point clouds of the same size is the sum of distances between a point in one cloud and a corresponding partner in the other minimized over all possible 1-to-1 correspondences. More formally,

\startcompact{small}
\begin{equation}
\lambda_{em} = \min_{\phi:A \rightarrow B}\sum_{a \in A} \Vert a - \phi(a)\Vert,
\label{eqn:em}
\end{equation}
\stopcompact{small}
where $\phi$ is a bijective mapping.

Point cloud metrics evaluated on vertices of 3D meshes can give misleading results, since large planar regions will have very few vertices, and hence contribute little. Instead, we evaluate these metrics using a
set of $n_s$ points $\mathbf{S}^{(i)} \in \mathbb{R}^{n_s \times 3}$ sampled uniformly from the surface of each 3D mesh.

\paragraph{\textbf{Intersection over union.}} As the name suggests, the intersection-over-union of volumetric representations $IoU$ is defined by the ratio of the volumes of the intersection over their union,
\startcompact{small}
\begin{equation}
\text{IoU}(A,B) = \dfrac{|A \cap B|}{|A \cup B|}.
\label{eqn:iou}
\end{equation}
\stopcompact{small}

\subsection{Deforming 3D Meshes}
We deform a 3D object by freely manipulating some control points using the \textsc{Ffd} technique. \textsc{Ffd} creates a grid of control points and its axes are defined by the orthogonal vectors $\mathbf{s}, \mathbf{t}$ and $\mathbf{u}$ \cite{Sederberg1986}. The control points are then defined by $l,m$ and $n$ which divides the grid in $l+1, m+1, n+1$ planes in the $\mathbf{s},\mathbf{t},\mathbf{u}$ directions, respectively. A local coordinate for each object's vertex is then imposed.

In this work, we deform an object through a trivariate Bernstein tensor as in \cite{Sederberg1986} which is basically a weighted sum of the control points. The deformed position of any arbitrary point is given by

\startcompact{small}
\begin{equation}
	\mathbf{s}(s,t,u) = \sum^l_{i=0} \sum^m_{j=0} \sum^n_{k=0}B_{i,l}(s)B_{j,m}(t)B_{k,n}(u)\mathbf{p}_{i,j,k},
 	\label{eq:FFD3}
\end{equation}
\stopcompact{small}

\noindent where $\mathbf{s}$ contains the coordinates of the displaced point, $B_{\theta,n}(x)$ is the Bernstein polynomial of degree $n$ which sets the influence of each control point on every model's vertex, and $\mathbf{p}_{i,j,k}$ is the $i,j,k$-th control point. Equation~\eqref{eq:FFD3} is a linear function of $\mathbf{P}$ and it can be written in a matrix form as
%
%

\startcompact{small}
\begin{equation}
	\mathbf{S} = \mathbf{BP},
 	\label{eq:FFD6}
\end{equation}
\stopcompact{small}

\noindent where the rows of $\mathbf{S} \in \mathbb{R}^{N \times 3}$ are the vertices of the 3D mesh, $\mathbf{B} \in \mathbb{R}^{N \times M}$ is the deformation matrix, $\mathbf{P} \in \mathbb{R}^{M \times 3}$ are the control point coordinates, and $N$ and $M$ are the number of vertices and control points, respectively.

\section{Learning Free-Form Deformations}
Our method involves applying deformations encoded with a parameter $\Delta \mathbf{P}^{(t)}$ to $T$ different template models $c^{(t)}$ with $0 \le t < T$. We begin by calculating the Bernstein decomposition of the face-sample point cloud of each template model, $\mathbf{S}^{(t)} = \mathbf{B}^{(t)} \mathbf{P}^{(t)}$. We use a CNN to map a rendered image to a set of shared high level image features. These features are then mapped to a grid deformation $\Delta \tilde{\mathbf{P}}^{(qt)}$ for each template independently to produce a perturbed set of surface points for each template,

\startcompact{small}
\begin{equation}
\tilde{\mathbf{S}}^{(qt)} = \mathbf{B}^{(t)}\left(\mathbf{P}^{(t)} + \Delta \tilde{\mathbf{P}}^{(qt)}\right).
\end{equation}
\stopcompact{small}

\noindent We also infer a weighting value $\gamma^{(qt)}$ for each template from the shared image features, and train the network using a weighted Chamfer loss,

\startcompact{small}
\begin{equation}
\lambda_0 = \sum_{q,\,t} f\left(\gamma^{(qt)}\right)\lambda_c\left(s^{(q)}, \tilde{s}^{(qt)}\right),
\label{eqn:cad-cnn}
\end{equation}
\stopcompact{small}
\noindent where $f$ is some positive monotonically increasing scalar function.

In this way, the network learns to assign high weights to templates which it has learned to deform well based on the input image, while the sub-networks for each template are not highly penalized for examples which are better suited to other templates.
We enforce a minimum weight by using
\startcompact{small}
\begin{equation}
\gamma^{(qt)} = \left(1 - \epsilon_\gamma\right)\gamma_0^{(qt)} + \frac{1}{T}\epsilon_\gamma,
\label{eqn:loss-weights}
\end{equation}
\stopcompact{small}
where $\gamma_0^{(qt)}$ is the result of a softmax function where summation is over the templates and $\epsilon_\gamma$ is a small constant threshold, $0 < \epsilon_\gamma \ll 1$.

For inference and evaluation, we use the template with the highest weight,
\startcompact{small}
\begin{align}
	t^* = \argmax_t\gamma^{(t)}, \\
	\tilde{c}^{(q)} = \{\tilde{\mathbf{S}}^{(qt^*)}, \mathbf{F}^{(t^*)}\}.
\end{align}
\stopcompact{small}

\noindent Key advantages of the architecture are as follows:
\begin{itemize}
	\item no 3D convolutions are involved, meaning the network scales well with increased resolution;
    \item no discretization occurs, allowing higher precision than voxel-based methods;
    \item the output $\Delta \tilde{\mathbf{P}}$ can be used to generate an arbitrarily dense point cloud -- not necessarily the same density as that used during training; and
    \item a mesh can be inferred by applying the deformation to the Bernstein decomposition of the \textit{vertices} while maintaining the same face connections.
\end{itemize}
Drawbacks include,
\begin{itemize}
	\item the network size scales linearly with the number of templates considered; and
    \item there is at this time no mechanism to explicitly encourage topological or semantic similarity.
\end{itemize}

\subsection{Diversity in Model Selection}
Preliminary experiments showed training using standard optimizers with an identity weighting function $f$ resulting in a small number of templates being selected frequently.
This is at least partially due to a positive feedback loop caused by the interaction between the weighting sub-network and the deformation sub-networks. If a particular template deformation sub-network performs particularly well initially, the weighting sub-network learns to assign increased weight to this template. This in turn affects the gradients which flow through the deformation sub-network, resulting in faster learning, improved performance and hence higher weight in subsequent batches. We experimented with a number of network modifications to reduce this.
%
%

\paragraph{\textbf{Non-linear Weighting.}}
One problem with the identity weighting scheme ($f(\gamma) = \gamma)$ is that there is no penalty for over-confidence. A well-trained network with a slight preference for one template over all the rest will be inclined to put all weight into that template. By using an $f$ with positive curvature, we discourage the network from making overly confident inferences. We experimented with an entropy-inspired weighting $f(\gamma) = -\log(1 - \gamma)$.

\paragraph{\textbf{Explicit Entropy Penalty.}}
Another approach is to penalize the lack of diversity directly by introducing an explicit entropy loss term,

\startcompact{small}
\begin{equation}
\lambda_e = \sum_t \bar{\gamma}_b^{(t)}\log\left(\bar{\gamma}_b^{(t)}\right),
\end{equation}
\stopcompact{small}
\noindent where $\bar{\gamma}^{(t)}$ is the weight value of template $t$ averaged over the batch. This encourages an even distribution over the batch but still allows confident estimates for the individual inferences. For these experiments, the network was trained with a linear combination of weighted Chamfer loss $\lambda_0$ and the entropy penalty,

\startcompact{small}
\begin{equation}
	\lambda_e' = \lambda_0 + \kappa_e \lambda_e.
\end{equation}
\stopcompact{small}

While a large entropy error term encourages all templates to be assigned weight and hence all subnetworks to learn, it also forces all subnetworks to try and learn all possible deformations. This works against the idea of specialization, where each subnetwork should learn to deform their template to match query models close to their template. To alleviate this, we anneal the entropy over time

\startcompact{small}
\begin{equation}
	\kappa_e = e^{-b/b_0}\kappa_{e0},
    \label{eqn:exponential-annealing}
\end{equation}
\stopcompact{small}

\noindent where $\kappa_0$ is the initial weighting, $b$ is the batch index and $b_0$ is some scaling factor.

\paragraph{\textbf{Deformation Regularization.}}
In order to encourage the network to select a template requiring minimal deformation, we introduce a deformation regularization term,

\startcompact{small}
\begin{equation}
	\lambda_{r} = \sum_{q, t} \gamma^{(qt)} |\Delta \tilde{\mathbf{P}}^{(qt)}|^2,
\end{equation}
\stopcompact{small}

\noindent where $|\cdot|^2$ is the squared 2-norm of the vectorized input.

Large regularization encourages a network to select the closest matching template, though punishes subnetworks for deforming their template a lot, even if the result is a better match to the query mesh. We combine this regularization term with the standard loss in a similar way to the entropy loss term,

\startcompact{small}
\begin{equation}
	\lambda_{r}' = \lambda_0 + \kappa_{r}\lambda_{r},
\end{equation}
\stopcompact{small}

\noindent where $\kappa_{r}$ is an exponentially annealed weighting with initial value $\kappa_{r0}$.



\subsection{Deformed Mesh Inference}
\begin{figure*}[!b]
\centering
{\fontfamily{ptm}\selectfont {\scriptsize \textbf{\hspace{-10pt} (a) \hspace{52pt} (b) \hspace{52pt} (c) \hspace{52pt} (d) \hspace{52pt} (e) }}}

\includegraphics[trim={1cm 1cm 1cm 0cm},clip,width=0.9\textwidth,keepaspectratio]{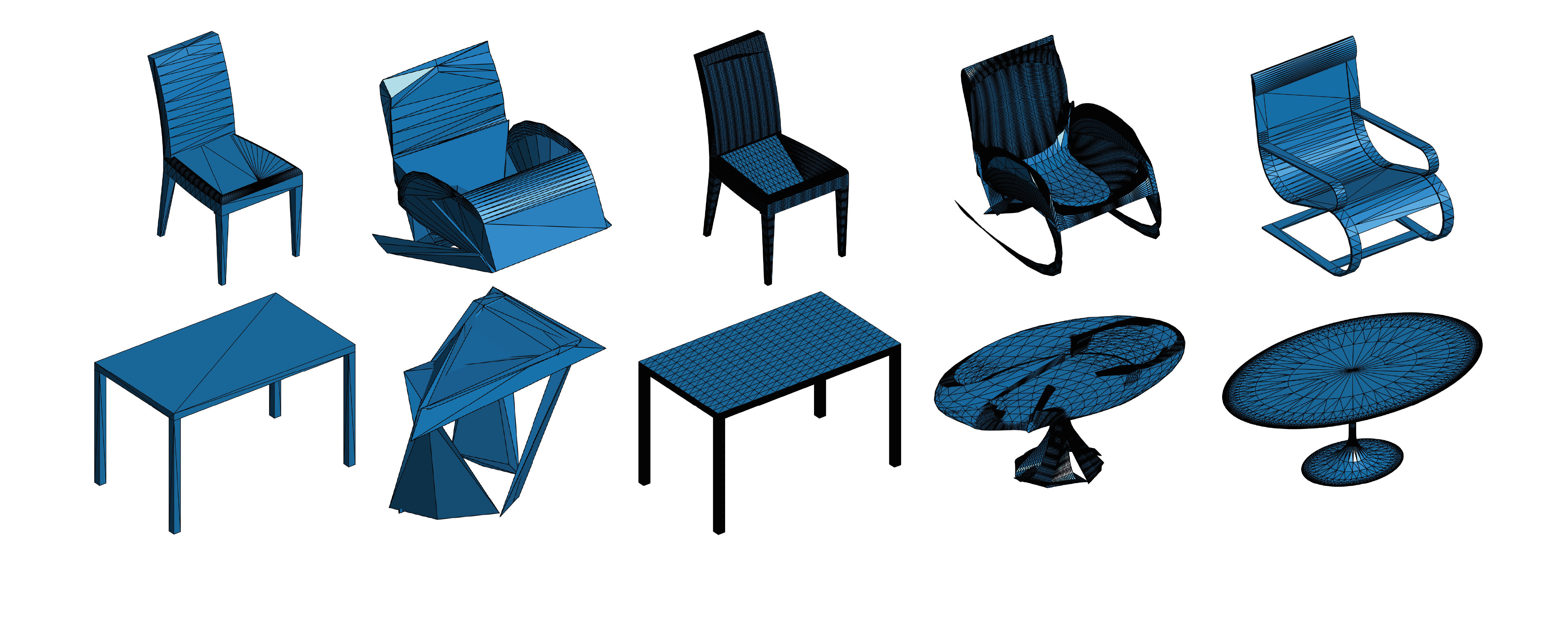}
\caption{Two examples of poor mesh model output (chair and table) as a result of low vertex density. (a) Original low vertex-density mesh. (b) Original mesh deformed according to inferred \textsc{Ffd}. (c) Subdivided mesh. (d) Subdivided mesh deformed according to same \textsc{Ffd}. (e) Ground truth.}
\label{fig:mesh-subdiv}
\end{figure*}

For the algorithm to result in high-quality 3D reconstructions,
it is important that the vertex density of each template mesh is approximately equivalent to (or higher than) the point cloud density used during training. To ensure this is the case, we subdivide edges in the template mesh such that no edge length is greater than some threshold $\epsilon_e$. Example cases where this is particularly important are illustrated in Figure \ref{fig:mesh-subdiv}.

\subsection{Implementation Details}
We employed a MobileNet architecture that uses depthwise separable convolutions to build light weight deep neural networks for mobile and embedded vision applications \cite{howard2017mobilenets} without the final fully connected layers and with width $\alpha = 0.25$. Weights were initialized from the convolutional layers of a network trained on the $192 \times 192$ ImageNet dataset \cite{ImageNet}. To reduce dimensionality, we add a single $1 \times 1$ convolution after the final MobileNet convolution layer. After flattening the result, we have one shared fully connected layer with $512$ nodes followed by a fully connected layer for each template. A summary of layers and output dimensions is given in Table \ref{tab:layer-dimensions}.

\begin{table}[ht!]
  \centering \small
  \resizebox{0.45\textwidth}{!}{
  \begin{tabular}{lc}
    \hline
    \textbf{Layer} \hspace{25mm} & \textbf{Output size} \\
    \hline
    Input image & $192 \times 256 \times 3$ \\
    MobileNet CNN & $6 \times 8 \times 256$ \\
    $1 \times 1$ convolution & $6 \times 8 \times 64$ \\
    Flattened & $3,072$ \\
    Shared FC & $512$ \\
    Template FC$\,^{(t)}$ & $192 + 1$\\
    \hline
  \end{tabular}}%
  \caption{Output size of network layers. Each template fully connected (FC) layer output is interpreted as $3(L+1)(M+1)(N+1) = 3 \times 4^3 = 192$ values for $\Delta \tilde{\mathbf{P}}^{(qt)}$ and a single $\gamma^{(qt)}$ value.}
  \label{tab:layer-dimensions}
\end{table}

We used a subset of the ShapeNet Core dataset \cite{ShapeNet} over a number of categories, using an 80/20 train/evaluation split. All experiments were conducted using $4$ control points in each dimension ($l = m = n = 3$) for the free form parametrizations. To balance computational cost with loss accuracy, we initially sampled all models surfaces with $16,384$ points for both labels and free form decomposition. At each step of training, we sub-sampled a different $1,024$ points for use in the Chamfer loss.

All input images were $192 \times 256 \times 3$ and were the result of rendering each textured mesh from the same view from $30^\circ$ above the horizontal, $45^\circ$ away from front-on and well-lit by a light above and on each side of the model. We trained a different network with 30 templates for each category. Templates were selected manually to ensure good variety. Models were trained using a standard Adam optimizer with learning rate $10^{-3}$, $\beta_1=0.9$, $\beta_2=0.999$ and $\epsilon=10^{-8}$. Mini-batches of 32 were used, and training was run for $b_{\text{max}}=100,000$ steps. Exponential annealing used $b_0 = 10,000$.

For each training regime, a different model was trained for each category. Hyper-parameters for specific training regimes are given in Table \ref{tab:hyper-params}.

\begin{table}
	\centering
    \resizebox{0.8\textwidth}{!}{
    \setlength{\tabcolsep}{8pt}
    \begin{tabular}{lccccc}
    \hline
    \textbf{Training Regime} & \textbf{ID} & $\epsilon_\gamma$ & $f(\gamma)$ & $\kappa_{e0}$ & $\kappa_{r0}$ \\
    \hline
    base & b & 0.1 & $\gamma$ & 0 & 0 \\
    log-weighted & w & 0.001 & $-\log(1 - \gamma)$ & 0 & 0 \\
    entropy & e & 0.1 & $\gamma$ & 100 & 0 \\
    regularized & r & 0.1 & $\gamma$ & 0 & 1 \\
    \hline
    \end{tabular}}
    \caption{Hyper parameters for the primary training regimes.}
	\label{tab:hyper-params}
\end{table}

To produce meshes and subsequent voxelizations and IoU scores, template meshes had edges subdivided to a maximum length of $\epsilon_e = 0.02$. We voxelize on a $32^3$ grid by initially assigning any voxel containing part of the mesh as occupied, and subsequently filling in any unoccupied voxels with no free path to the outside.

\section{Experimental Results}
\begin{figure}[ht!]
\centering
{\fontfamily{ptm}\selectfont {\scriptsize \textbf{\hspace{-2pt} (a) Input \hspace{20pt} (b) Template \hspace{5pt} (c) Deformed PC \hspace{5pt} (d) Deformed Voxels \hspace{5pt} (e) Deformed mesh \hspace{5pt} (f) GT}}}
\includegraphics[trim={0.3cm 0cm 0cm 0cm},clip,width=1.04\textwidth,keepaspectratio]{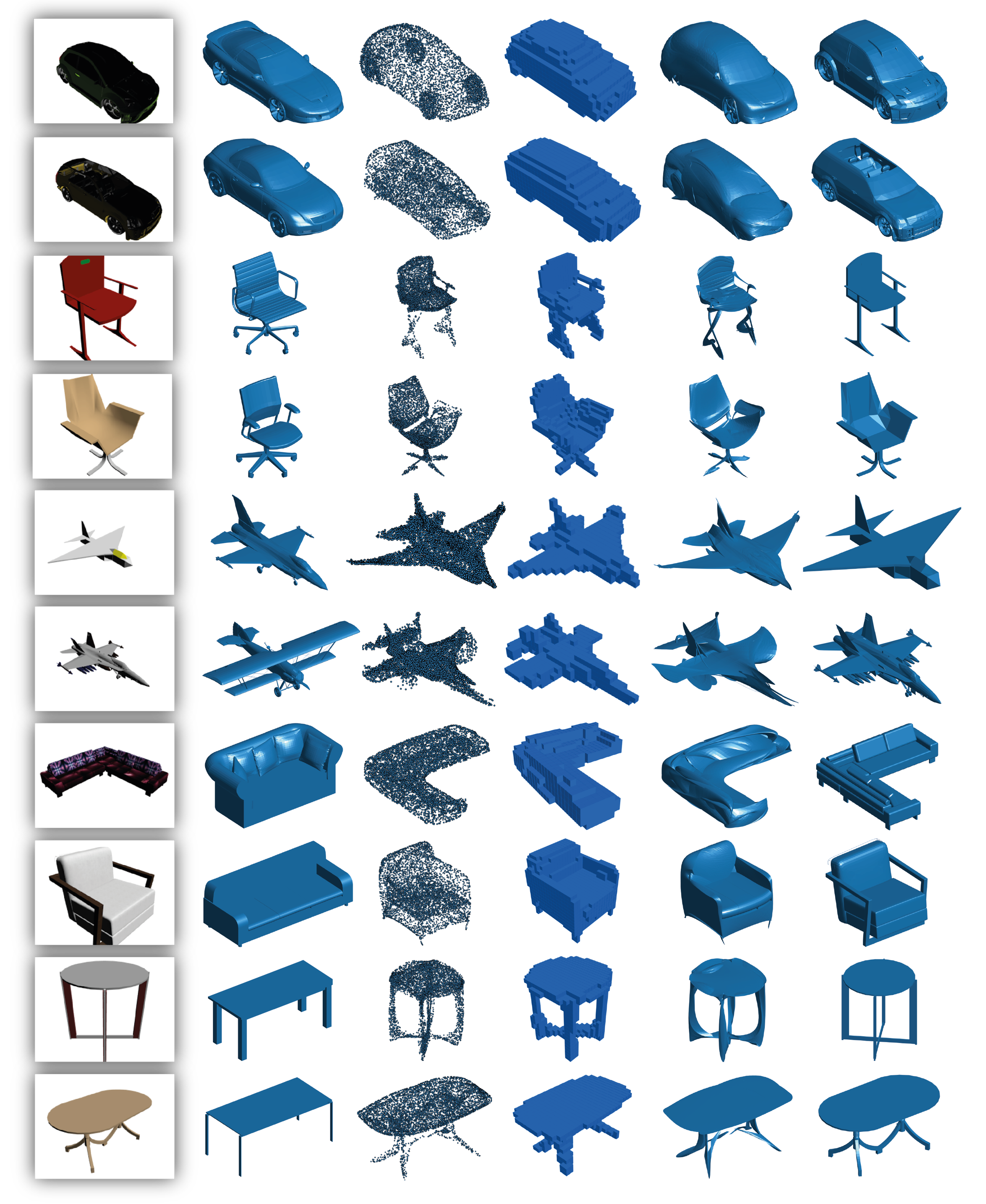}
\caption{Representative results for different categories. Column (a) shows the input image. Column (b) shows the selected template model. Column (c) shows the deformed point cloud by \textsc{Ffd}. The deformed voxelized model is shown in column (d). Column (e) shows our final 3D mesh reconstruction, and the ground truth is shown in column (f).}
\label{fig:results-normal}
\end{figure}

Qualitatively, we observe the networks preference in applying relatively large deformations to geometrically simple templates, and do not shy away from merging separate features of template models. For example, models frequently select the bi-plane template and merge the wings together to match single-wing aircraft, or warp standard 4-legged chairs and tables into structurally different objects as shown in Figure \ref{fig:results-normal}.

\subsection{Quantitative Comparison}
For point cloud comparison, we compare against the works of Kuryenkov \etal \cite{kuryenkov2017deformnet} (DN) and Fan \etal \cite{fan2016point} (PSGN) for 5 categories. We use the results for the improved PSGN model reported in \cite{kuryenkov2017deformnet}. We use the same scaling as in these papers, finding transformation parameters that transform the ground-truth meshes to a minimal bounding hemisphere $z \ge 0$ of radius $3.2$ and applying this transformation to the inferred clouds. We also compare IoU values with PSGN \cite{fan2016point} on an additional 8 categories for voxelized inputs on a $32^3$ grid. Results for all 13 categories with each different training regime are given in Table \ref{tab:results}.

\begin{table}[b!]
\centering
\setlength{\tabcolsep}{7pt}
\resizebox{0.91\textwidth}{!}{\begin{tabular}{@{}rccccc|cc@{}}
\toprule
&\multicolumn{4}{c}{\textbf{Training Regime}} &\hphantom &\multicolumn{2}{c}{\textbf{Other Methods}} \\
\cmidrule{2-5} \cmidrule{7-8}
&b &w &e &r &&DN &PSGN \\ \cmidrule(r{1pt}){2-2} \cmidrule(r{1pt}){3-3} \cmidrule(r{1pt}){4-4} \cmidrule(r{1pt}){5-5} \cmidrule(r{1pt}){7-7} \cmidrule(r{1pt}){8-8}

\textbf{plane} 			&31/306/292 	&31/310/297 	&31/300/289 	&33/304/307		&&100/560/- &140/115/399	 \\
\textbf{bench} 			&42/280/431 	&39/280/425 	&40/284/418 	&45/275/445 	&&100/550/- &210/980/450 	 \\
\textbf{car}			&58/328/210 	&60/333/219 	&58/325/207 	&59/324/216 	&&90/520/-	&110/380/169 	 \\
\textbf{chair}			&36/280/407 	&35/275/393 	&35/277/392 	&37/277/401 	&&130/510/- &330/770/456 	 \\
\textbf{sofa}			&64/329/275 	&63/320/275 	&64/324/271 	&65/319/276 	&&210/770/- &230/600/292 	 \\ \midrule
\textbf{mean$_5$}		&\textbf{46}/305/323 	&\textbf{46}/304/322 	&\textbf{46}/300/\textbf{315} 	&48/\textbf{292}/329		&&130/580/- &200/780/353 	 \\ \midrule
\textbf{cabinet}		&37/249/282		&37/250/282 	&36/251/264 	&37/246/279 	&&- 	&-/-/229 	 \\
\textbf{monitor}		&38/253/369 	&37/250/367 	&37/255/367 	&43/255/380 	&&- 	&-/-/448 	 \\
\textbf{lamp}			&52/402/514 	&49/393/480 	&44/384/473 	&55/425/520 	&&-		&-/-/538 	 \\
\textbf{speaker}		&72/312/301 	&68/309/304 	&71/313/301 	&73/308/315 	&&- 	&-/-/263 	 \\
\textbf{firearm}		&39/312/332 	&30/279/281		&32/288/326 	&39/301/345 	&&- 	&-/-/396 	 \\
\textbf{table}			&47/352/447 	&46/331/432		&46/342/420 	&49/319/450 	&&- 	&-/-/394 	 \\
\textbf{cellphone}		&16/159/241 	&15/150/224		&15/154/192 	&15/154/222 	&&- 	&-/-/251 	 \\
\textbf{watercraft}		&83/408/493 	&48/296/340		&49/304/361 	&53/317/367 	&&- 	&-/-/389 	 \\	\midrule
\textbf{mean$_{13}$}	&47/305/353 		&\textbf{43}/\textbf{290}/332		&\textbf{43}/292/\textbf{329} 	&46/294/348 	&&- 	&250/800/360 	 \\

\bottomrule
\end{tabular}}
\caption{$1000\times (\lambda_c$/$\lambda_{em}$/$1 - \text{IoU})$ for our different training regimes, compared against state-of-the-art models DN \cite{kuryenkov2017deformnet} and PSGN \cite{fan2016point}. Lower is better. $\lambda_c$ and $\lambda_{em}$ values for PSGN are from the latest version as reported by Kuryenkov \etal \cite{kuryenkov2017deformnet}, while IoU values are from the original paper. $\text{mean}_5$ is the mean value across the plane, bench, car, chair and sofa categories, while $\text{mean}_{13}$ is the average across all 13.}
\label{tab:results}
\end{table}

\begin{figure*}[t!]
\centering
  \subfloat[]{
    \centering
    \includegraphics[scale=0.22]{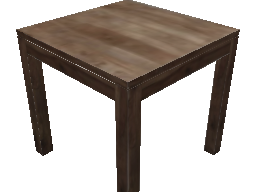}
   }
   \hspace{4mm}
   \qquad
  \subfloat[]{
    \centering
    \includegraphics[scale=0.13]{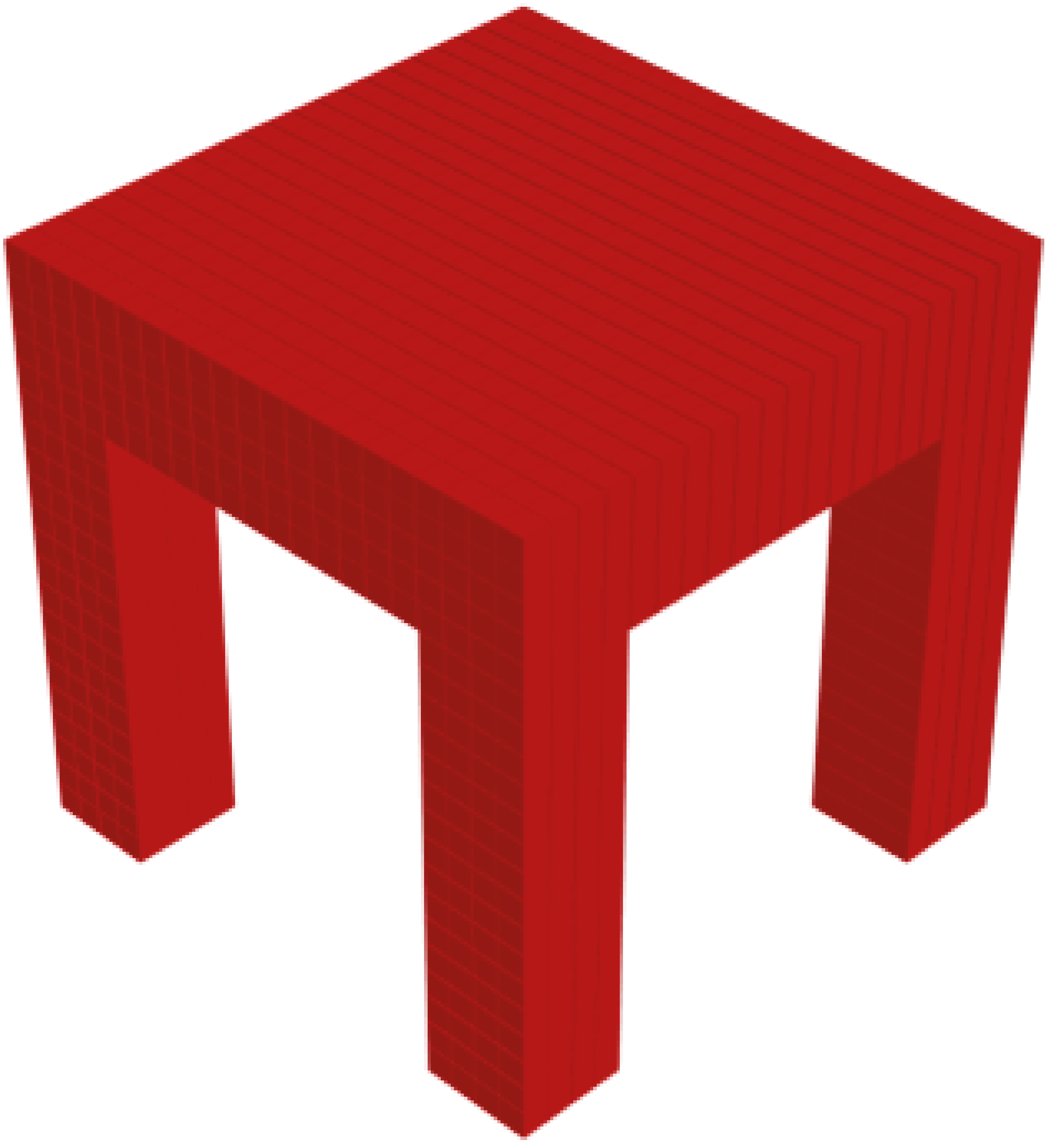}
   }
    \hspace{4mm}
	\qquad
  \subfloat[]{
    \centering
    \includegraphics[scale=0.14]{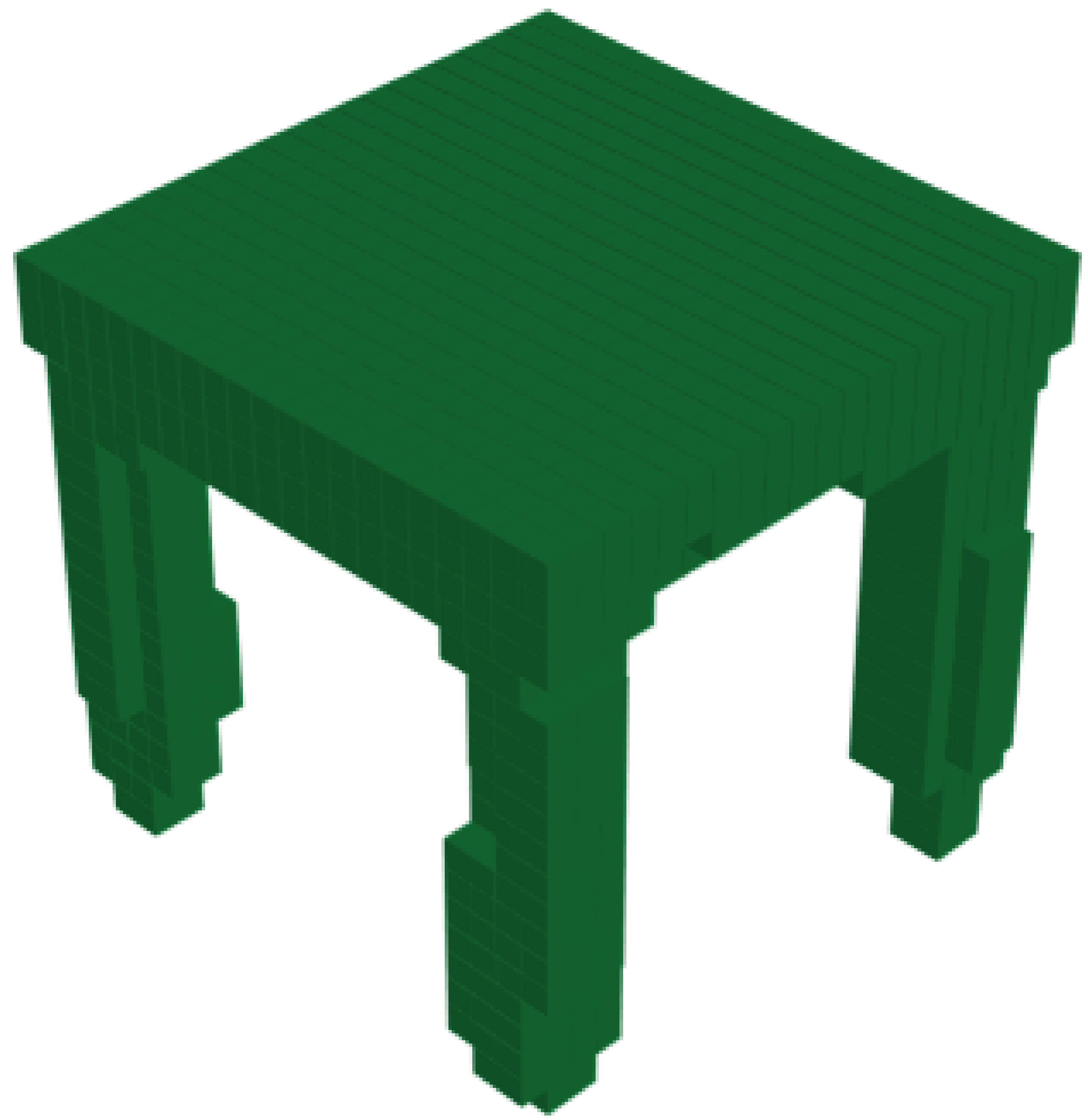}
  }
   \hspace{4mm}
  \qquad
  \subfloat[]{
    \centering
    \includegraphics[scale=0.13]{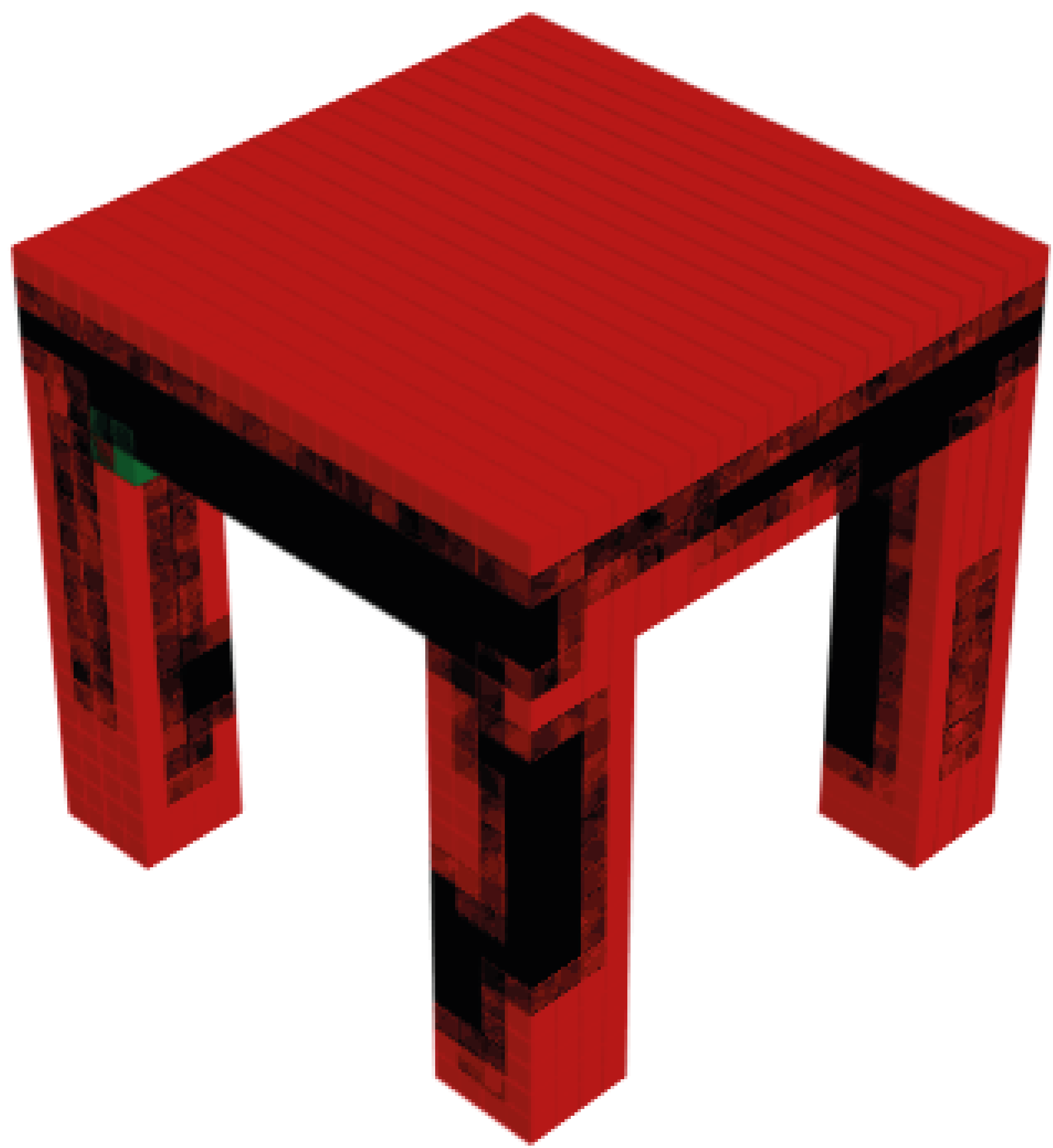}
  }
	\caption{An example of a qualitatively accurate inference with a low IoU score, $\lambda_{IoU} = 0.33$. Small errors in depth/width inference correspond to small Chamfer losses. For comparison, black voxels are true positives (intersection), red voxels are false negatives and green voxels are false positives.}
    \label{fig:table-error}
\end{figure*}

All our training regimes out-perform the others by a significant margin on all categories for point-cloud metrics ($\lambda_c$ and $\lambda_{em}$). We also outperform PSGN on IoU for most categories and on average. The categories for which the method performs worst in terms of IoU -- tables and chairs -- typically feature large, flat surfaces and thin structures. Poor IoU scores can largely be attributed to poor width or depth inference (a difficult problem given the single view provided) and small, consistent offsets that do not induce large Chamfer losses. An example is shown in Figure \ref{fig:table-error}.

\subsection{Template Selection}
\begin{figure}[!ht]
   \centering
   \subfloat[][]{\includegraphics[trim={0.5cm 0.1cm 1cm 1cm},clip,width=.25\textwidth]{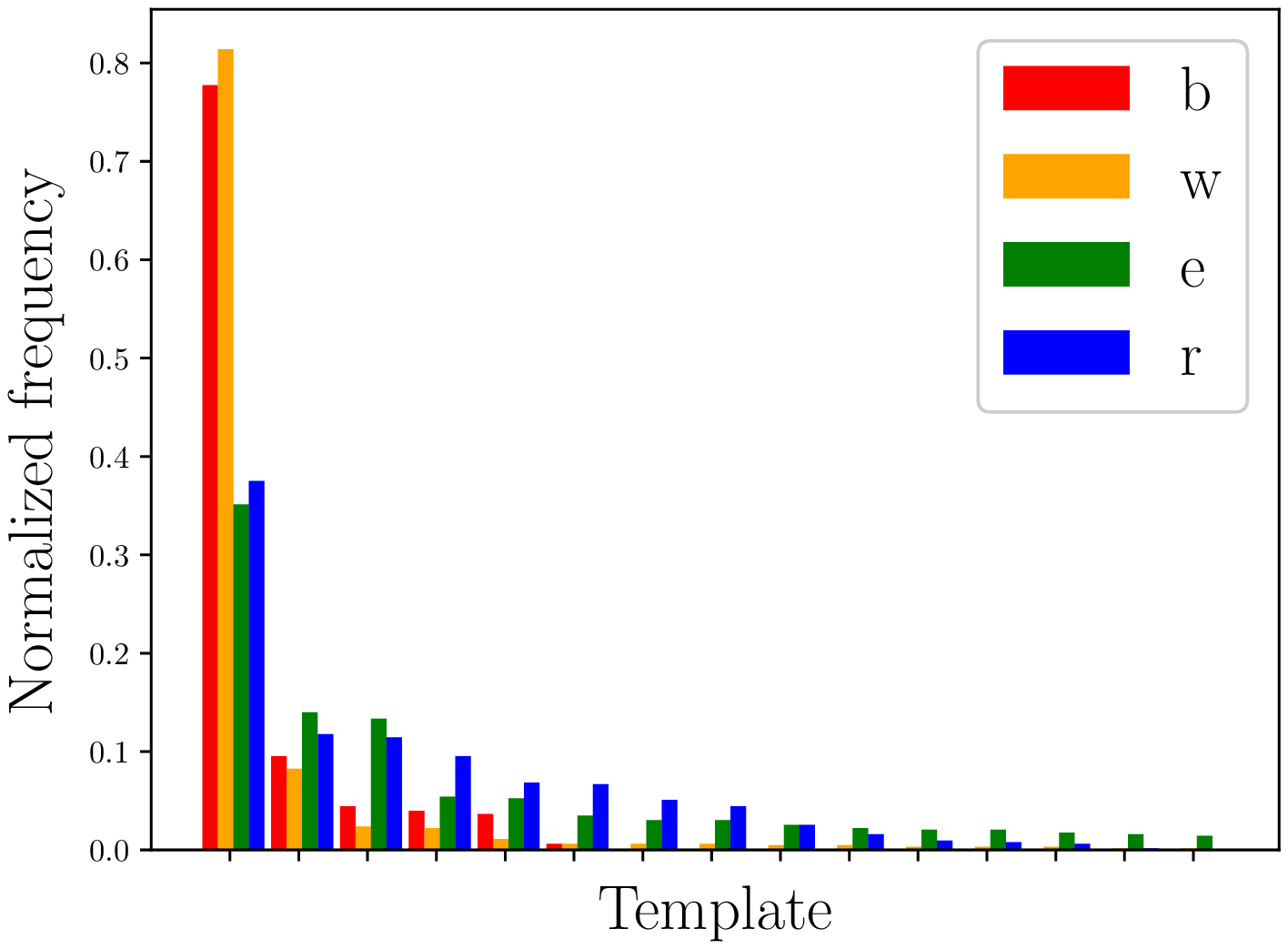}}
   \subfloat[][]{\includegraphics[trim={0.35cm 0.1cm 0.8cm 1cm},clip,width=.256\textwidth]{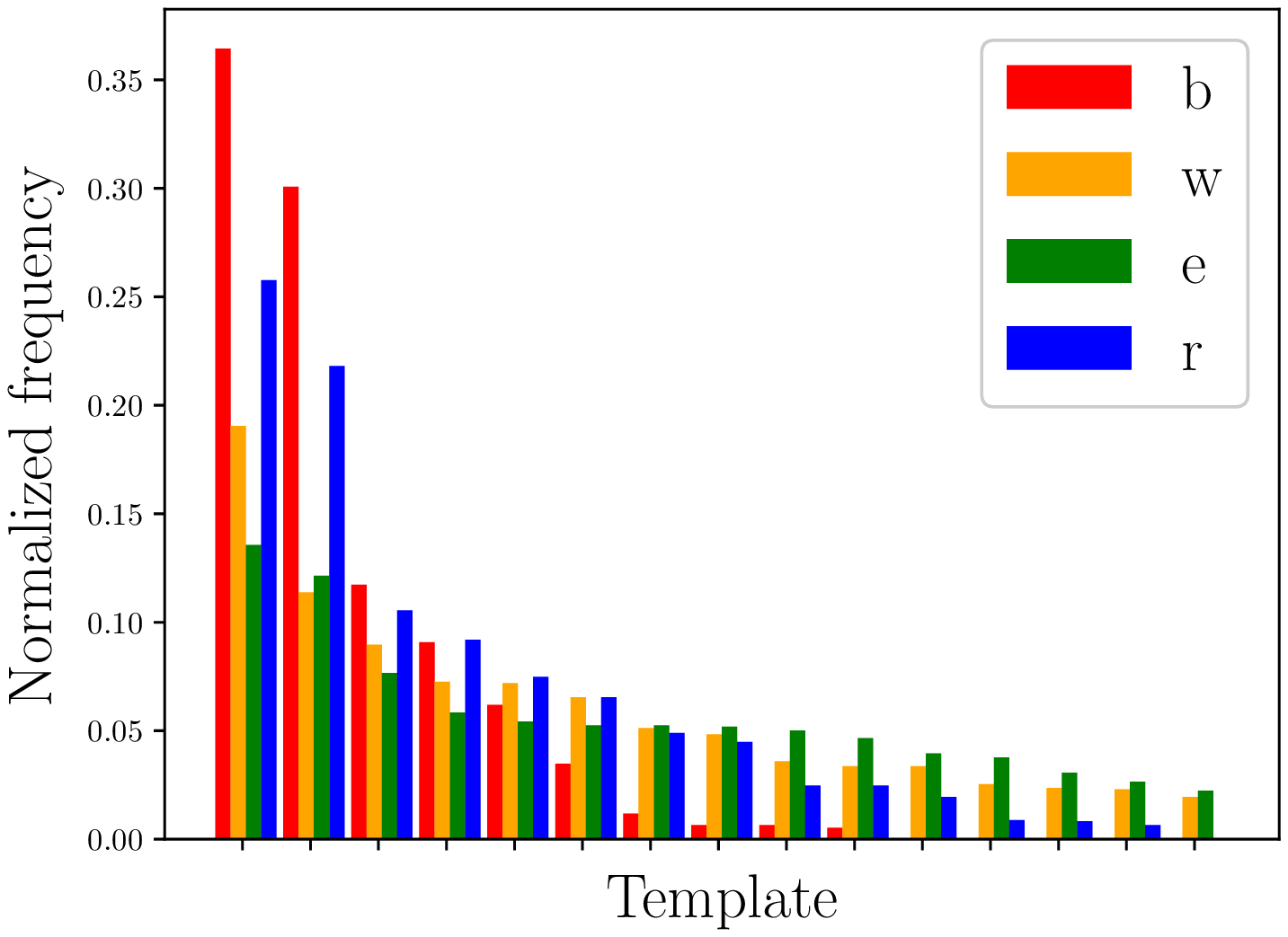}}
   \subfloat[][]{\includegraphics[trim={0.5cm 0.1cm 0.8cm 1cm},clip,width=.254\textwidth]{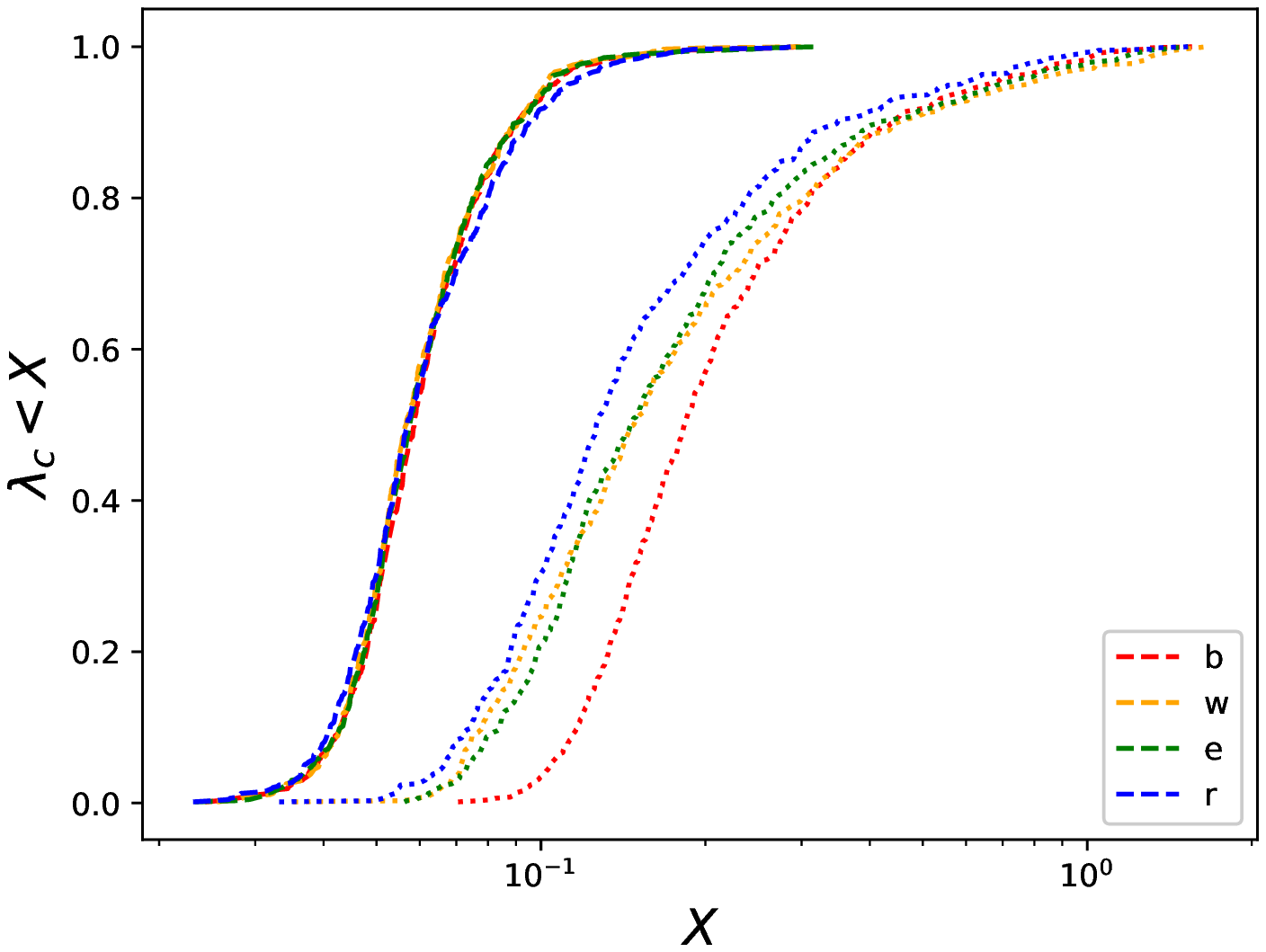}}
   \subfloat[][]{\includegraphics[trim={0.5cm 0.1cm 0.8cm 1cm},clip,width=.254\textwidth]{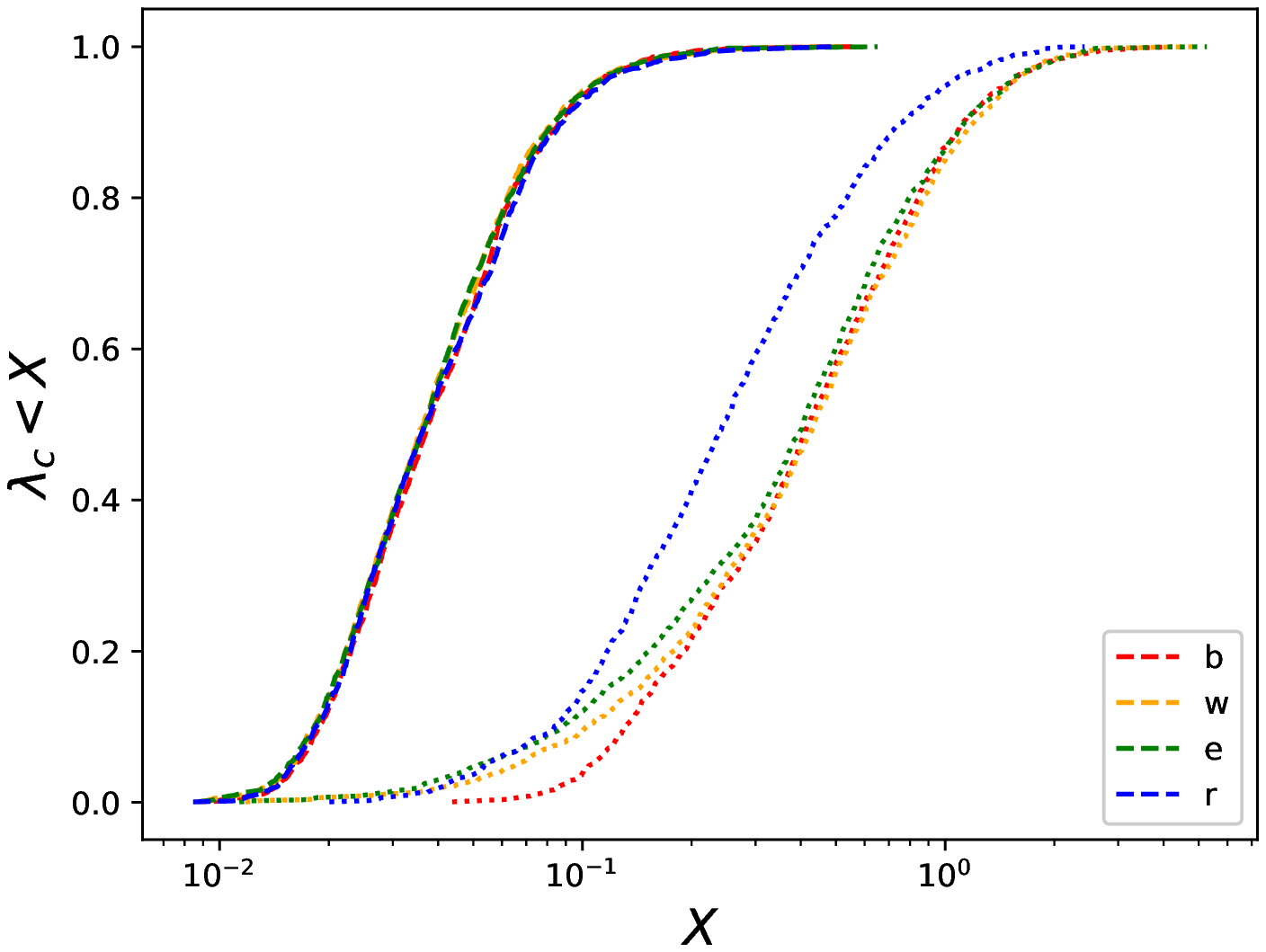}}
   \caption{Normalized count of the number of times each template was selected, sorted descending (a,b), and cumulative Chamfer error (c,d) for the deformed models (dashed) and undeformed template models (dotted) for the sofa category (a,c) and table category (b,d).
      \label{fig:model-result-plots}}
\end{figure}

We begin our analysis by investigating the number of times each template was selected across the different training regimes, and the quality of the match of the undeformed template to the query model. Results for the sofa and table categories are given in Figure \ref{fig:model-result-plots}.

We illustrate the typical behaviour of our framework with the sofa and table categories, since these are categories with topologically similar models and topologically different models respectively. In both cases, the base training regime (b) resulted in a model with template selection dominated by a small number of templates, while additional loss terms in the form of deformation regularization (r) and entropy (e) succeeded in smearing out this distribution to some extent. The behaviour of the non-linear weighting regime (w) is starkly different across the two categories however, reinforcing template dominance for the category with less topological differences across the dataset, and encouraging variety for the table category.

In terms of the Chamfer loss, all training regimes produced deformed models with virtually equivalent results. The difference is apparent when inspecting the undeformed models. Unsurprisingly, penalizing large deformation via regularization results in the best results for the undeformed template, while the other two non-base methods selected templates slightly better than the base regime.

\begin{figure}[!b]
  \centering
  \subfloat[][]{
    \centering
    \includegraphics[trim={0.5cm 0.1cm 0.8cm 1cm},clip,scale=0.28]{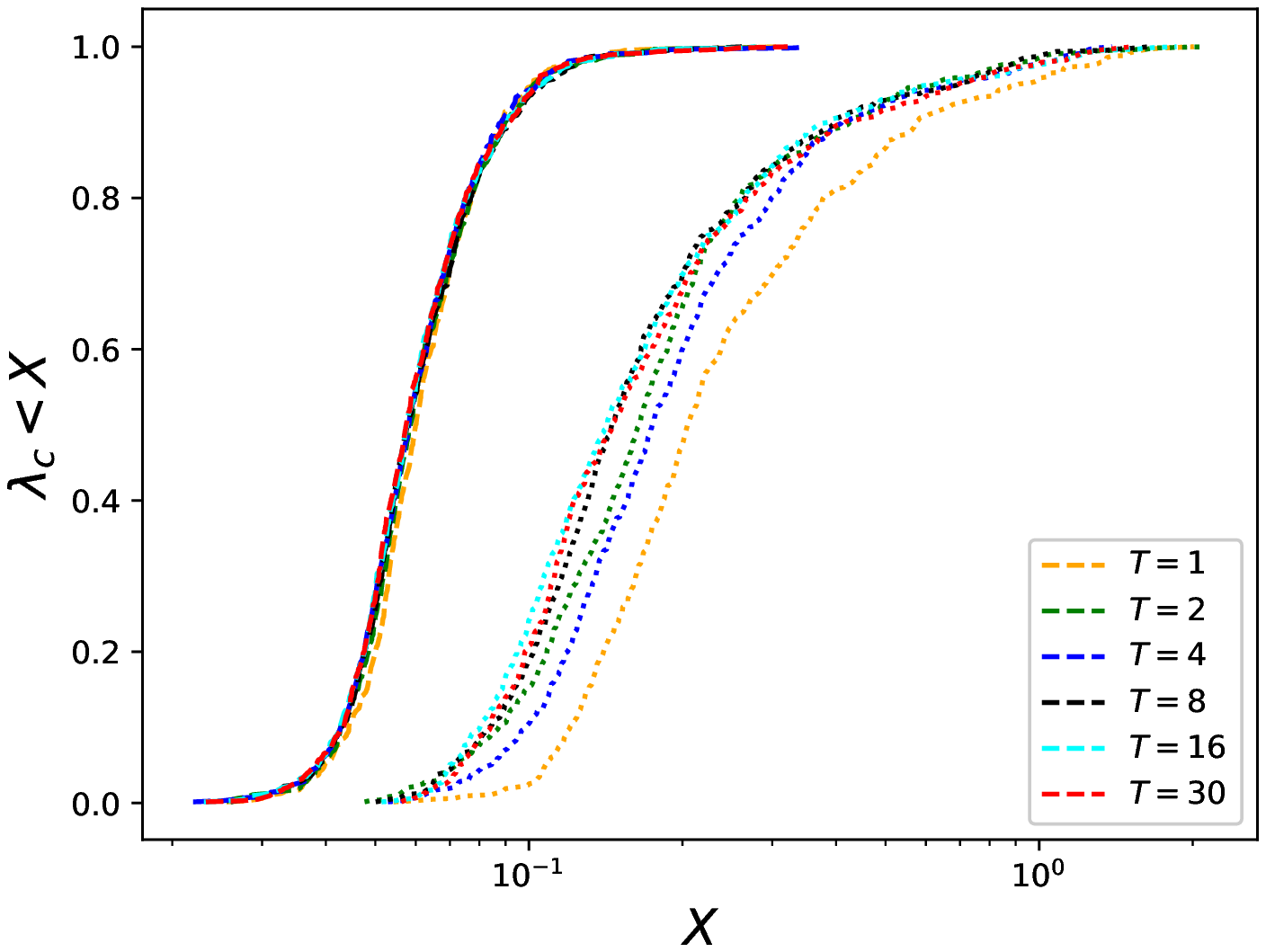}
   }
   \hspace{10mm}
  \subfloat[][]{
    \centering
    \includegraphics[trim={0.5cm 0.1cm 0.8cm 1cm},clip,scale=0.28]{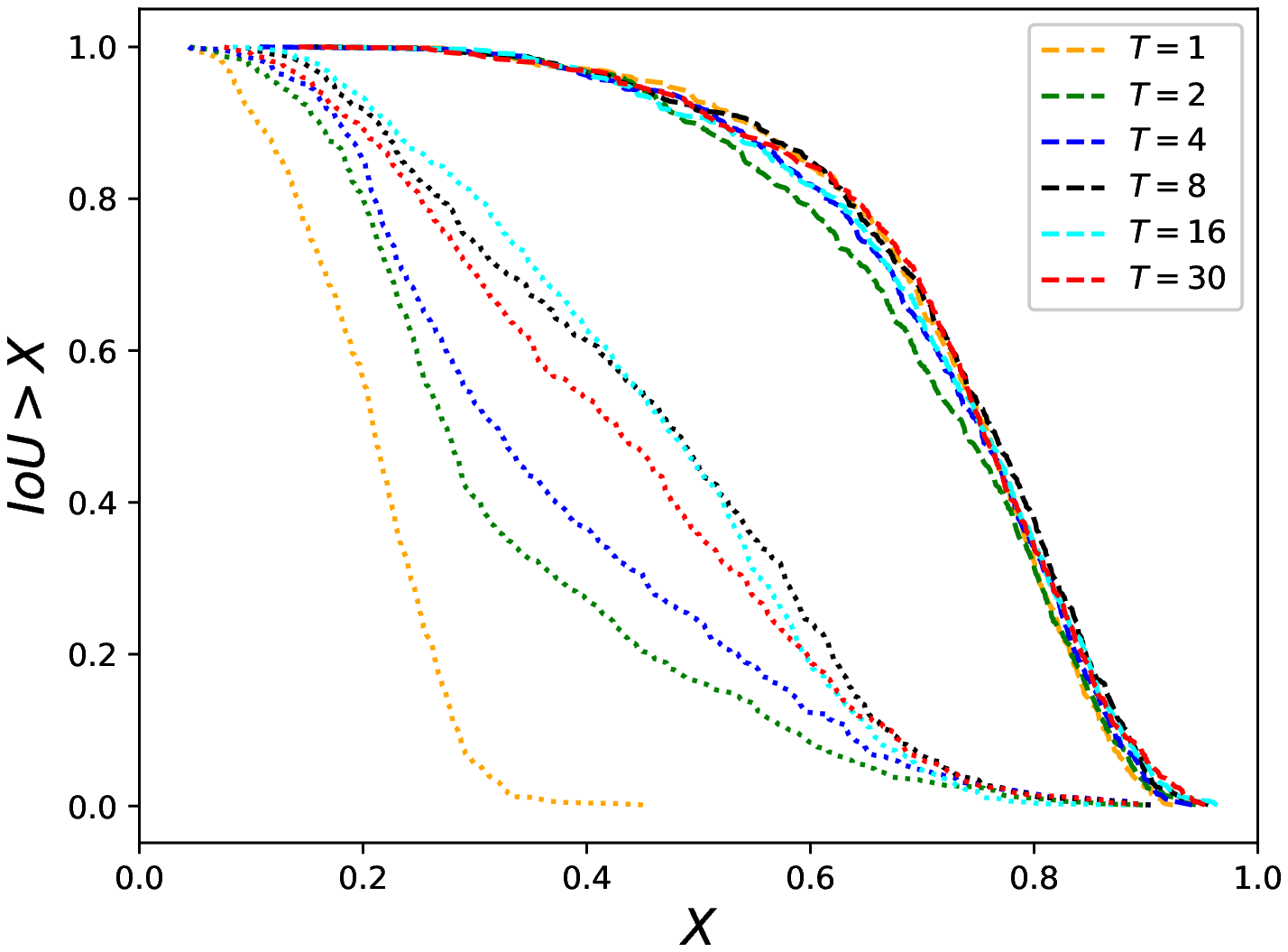}
  }
  \caption{Cumulative Chamfer loss (left) and IoU results (right) for models with limited templates. All models with $T > 1$ trained under the entropy regime (e) on the sofa category. $T=1$ model was trained with the base training regime, b. Values for the undeformed selected template are dotted, while deformed model values are dashed.
  \label{fig:limited-template-plots}}
\end{figure}

To further investigate the effect of template selection on the model, we trained a base model with a single template ($T = 1$), and entropy models with $T \in \{2, 4, 8, 16\}$ templates for the sofa dataset. In each case, the top $N$ templates selected by the 30-template regularized model were used. Cumulative Chamfer losses and IoU scores are shown in Figure \ref{fig:limited-template-plots}.

Surprisingly, the deformation networks manage to achieve almost identical results on these metrics regardless of the number of templates available. Additional templates do improve accuracy of the undeformed model up to a point, suggesting the template selection mechanism is not fundamentally broken.

%
%

\subsection{Semantic Label Transfer}
\begin{figure}[!ht]
   \centering
   \subfloat{\includegraphics[trim={1cm 0.3cm 1cm 0cm},clip,width=.15\textwidth]{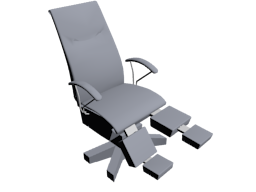}}
   \subfloat{\includegraphics[trim={4cm 1cm 4cm 1cm},clip,width=.10\textwidth]{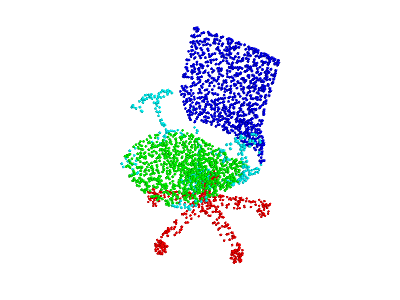}}
   \subfloat{\includegraphics[trim={4cm 1cm 4cm 1cm},clip,width=.10\textwidth]{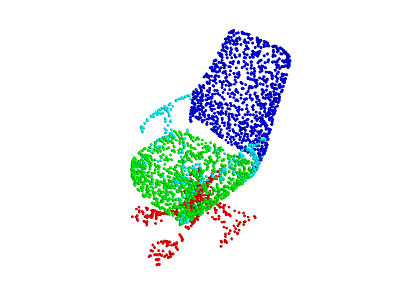}}
   \subfloat{\includegraphics[trim={4cm 1cm 4cm 1cm},clip,width=.10\textwidth]{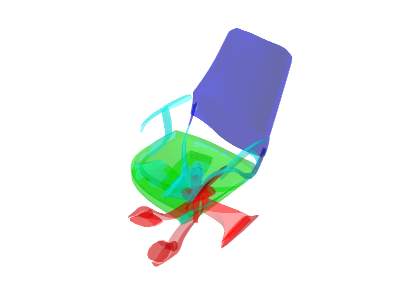}} \qquad
   \subfloat{\includegraphics[trim={1cm 0cm 1cm 0cm},clip,width=.13\textwidth]{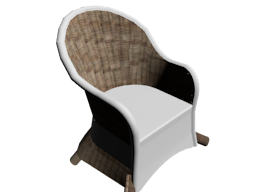}} \;
   \subfloat{\includegraphics[trim={4cm 0.5cm 4cm 1cm},clip,width=.10\textwidth]{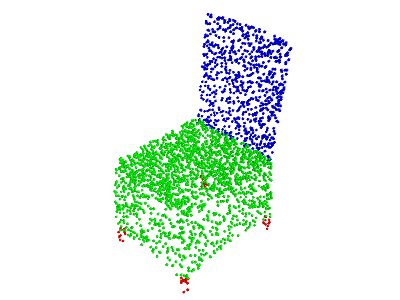}} \;
   \subfloat{\includegraphics[trim={4cm 0.5cm 4cm 1cm},clip,width=.10\textwidth]{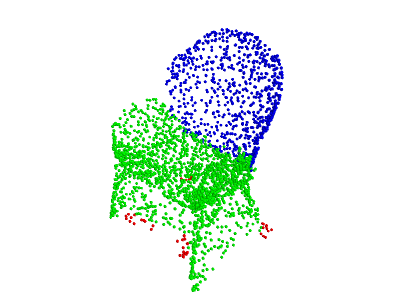}} \;
   \subfloat{\includegraphics[trim={4cm 0.5cm 4cm 1cm},clip,width=.10\textwidth]{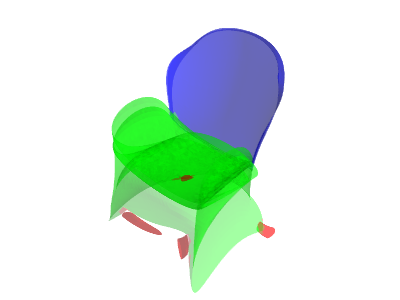}} \\
   \vspace{-0.4cm}
   \subfloat{\includegraphics[trim={1cm 0cm 1cm 1.5cm},clip,width=.15\textwidth]{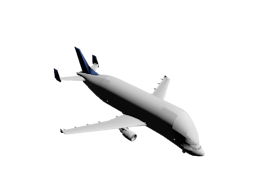}}
   \subfloat{\includegraphics[trim={4cm 0.5cm 4cm 2.5cm},clip,width=.1\textwidth]{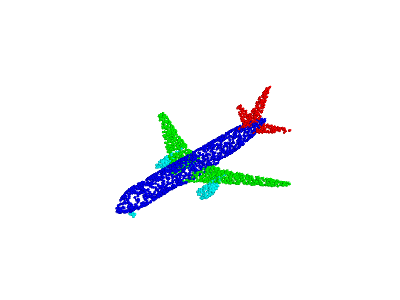}}
   \subfloat{\includegraphics[trim={4cm 0.5cm 4cm 2.5cm},clip,width=.1\textwidth]{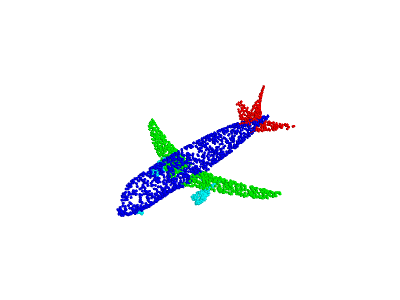}}
   \subfloat{\includegraphics[trim={4cm 0.5cm 4cm 2.5cm},clip,width=.1\textwidth]{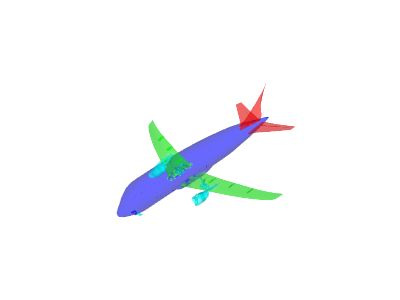}} \qquad
   \subfloat{\includegraphics[trim={1cm 0cm 1cm 1cm},clip,width=.15\textwidth]{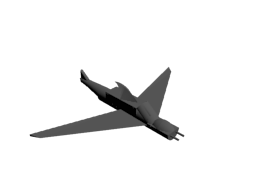}}
   \subfloat{\includegraphics[trim={4cm 1cm 4cm 2.5cm},clip,width=.1\textwidth]{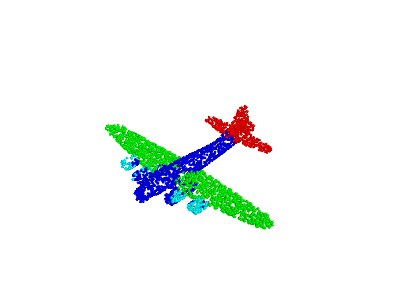}}
   \subfloat{\includegraphics[trim={4cm 2cm 4cm 2.5cm},clip,width=.12\textwidth]{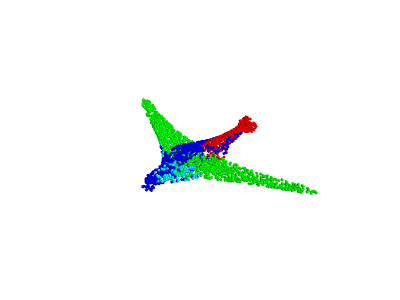}}
   \subfloat{\includegraphics[trim={4cm 2cm 4cm 2.5cm},clip,width=.12\textwidth]{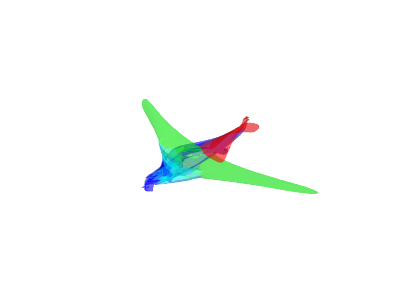}}
   \caption{Good (left block) and bad (right block) examples from the chair and plane categories. For each block: Input image; selected template's semantically segmented cloud; deformed segmented cloud; deformed mesh. Models trained with additional entropy loss term (e). 
   }
   \label{fig:segmented-clouds}
\end{figure}

While no aspect of the training related to semantic information, applying the inferred deformations to a semantically labelled point cloud allows us to infer another semantically labelled point cloud. Some examples are shown in Figure \ref{fig:segmented-clouds}. For cases where the template is semantically similar to the query object, the additional semantic information is retained in the inferred cloud. However, some templates either do not have points of all segmentation types, or have points of segmentation types that are not present in the query object. In these cases, while the inferred point cloud matches the surface relatively well, the semantic information is unreliable.

\section{Conclusion}
We have presented a simple framework for combining modern CNN approaches with detailed, unstructured meshes by using \textsc{Ffd} as a fixed sized intermediary and simultaneously learning to select and deform template point clouds based on minimally adjusted off-the-shelf image processing networks. We significantly out-perform state-of-the-art methods with respect to point cloud generation, and perform at-or-above state-of-the-art on the volumetric IoU metric, despite our network not being optimized for it. We present various mechanisms by which the diversity of templates selected can be increased and demonstrate these result in modest improvements.

We demonstrate the main component of the low metric scores is the ability of the network to learn deformations tailored to specific templates, rather than the precise selection of these templates. Models with only a single template to select from achieve comparable results to those with a greater selection at their disposal. This indicates the choice of template -- and hence any semantic of topological information -- makes little difference to the resulting point cloud, diminishing the trustworthiness of such topological or semantic information about the inferred model.


\clearpage

\section*{Acknowledgements}
This research was supported by the ARC grants DP170100632 and FT170100072. Computational resources and services used in this work were provided by the HPC and Research Support Group, Queensland University of Technology, Brisbane, Australia.

\bibliographystyle{splncs}
\bibliography{egbib}

\begin{thebibliography}{10}

\bibitem{Penner2017soft3d}
Penner, E., Zhang, L.:
\newblock Soft {3D} reconstruction for view synthesis.
\newblock In: ACM Transactions on Graphics. Volume~36.
\newblock (2017)

\bibitem{huang2015}
Huang, Q., Wang, H., Koltun, V.:
\newblock Single-view reconstruction via joint analysis of image and shape
  collections.
\newblock In: ACM Transactions on Graphics. Volume~34.
\newblock (2015)

\bibitem{maier2017intrinsic3d}
Maier, R., Kim, K., Cremers, D., Kautz, J., Nie{\ss}ner, M.:
\newblock Intrinsic{3D}: High-quality {3D} reconstruction by joint appearance
  and geometry optimization with spatially-varying lighting.
\newblock In: ICCV. (2017)

\bibitem{NIPS2012_4824}
Krizhevsky, A., Sutskever, I., Hinton, G.E.:
\newblock Image{N}et classification with deep convolutional neural networks.
\newblock In: NIPS.
\newblock (2012)  1097--1105

\bibitem{farabet2013learning}
Farabet, C., Couprie, C., Najman, L., LeCun, Y.:
\newblock Learning hierarchical features for scene labeling.
\newblock TPAMI \textbf{35}(8) (2013)  1915--1929

\bibitem{graves2009novel}
Graves, A., Liwicki, M., Fern{\'a}ndez, S., Bertolami, R., Bunke, H.,
  Schmidhuber, J.:
\newblock A novel connectionist system for unconstrained handwriting
  recognition.
\newblock TPAMI \textbf{31}(5) (2009)  855--68

\bibitem{choy2016}
Choy, C.B., Xu, D., Gwak, J., Chen, K., Savarese, S.:
\newblock {3D-R2N2}: {A} unified approach for single and multi-view {3D} object
  reconstruction.
\newblock In: ECCV. (2016)

\bibitem{Yan2016}
Yan, X., Yang, J., Yumer, E., Guo, Y., Lee, H.:
\newblock Perspective transformer nets: Learning single-view {3D} object
  reconstruction without {3D} supervision.
\newblock In: NIPS. (2016)

\bibitem{QiSu2016}
Qi, C.R., Su, H., Nie{\ss}ner, M., Dai, A., Yan, M., Guibas, L.J.:
\newblock Volumetric and multi-view {CNNs} for object classification on {3D}
  data.
\newblock In: CVPR. (2016)

\bibitem{Kar2017}
Kar, A., H\"ane, C., Malik, J.:
\newblock Learning a multi-view stereo machine.
\newblock In: NIPS. (2017)

\bibitem{Zhu2017}
Zhu, R., Galoogahi, H.K., Wang, C., Lucey, S.:
\newblock Rethinking reprojection: {C}losing the loop for pose-aware shape
  reconstruction from a single image.
\newblock In: NIPS. (2017)

\bibitem{marrnet2017}
Wu, J., Wang, Y., Xue, T., Sun, X., Freeman, W.T., Tenenbaum, J.B.:
\newblock {MarrNet: 3D Shape Reconstruction via 2.5D Sketches}.
\newblock In: NIPS. (2017)

\bibitem{fan2016point}
Fan, H., Su, H., Guibas, L.J.:
\newblock A point set generation network for {3D} object reconstruction from a
  single image.
\newblock In: CVPR. (2017)

\bibitem{Qi_CVPR2017}
Qi, C.R., Su, H., Mo, K., Guibas, L.J.:
\newblock {PointNet:} {D}eep learning on point sets for {3D} classification and
  segmentation.
\newblock In: CVPR. (2017)

\bibitem{Qi2017}
Qi, C.R., Yi, L., Su, H., Guibas, L.J.:
\newblock {PointNet++:} {D}eep hierarchical feature learning on point sets in a
  metric space.
\newblock In: NIPS. (2017)

\bibitem{lin2017learning}
Lin, C.H., Kong, C., Lucey, S.:
\newblock Learning efficient point cloud generation for dense {3D} object
  reconstruction.
\newblock In: AAAI. (2018)

\bibitem{Sederberg1986}
Sederberg, T., Parry, S.:
\newblock Free-form deformation of solid geometric models.
\newblock In: SIGGRAPH. (1986)

\bibitem{Ulusoy2015}
Ulusoy, A.O., Geiger, A., Black, M.J.:
\newblock Towards probabilistic volumetric reconstruction using ray potential.
\newblock In: 3DV. (2015)

\bibitem{Wu2015}
Wu, Z., Song, S., Khosla, A., Tang, X., Xiao, J.:
\newblock {3D} {ShapeNets}: {A} deep representation for volumetric shapes.
\newblock In: CVPR. (2015)

\bibitem{Cherabier2016}
Cherabier, I., Häne, C., Oswald, M.R., Pollefeys, M.:
\newblock Multi-label semantic {3D} reconstruction using voxel blocks.
\newblock In: 3DV. (2016)

\bibitem{Sharma2016}
Sharma, A., Grau, O., Fritz, M.:
\newblock {VConv-DAE}: {D}eep volumetric shape learning without object labels.
\newblock In: ECCVW. (2016)

\bibitem{Rezende2016}
J.~Rezende, D., Eslami, S.M.A., Mohamed, S., Battaglia, P., Jaderberg, M.,
  Heess, N.:
\newblock Unsupervised learning of {3D} structure from images.
\newblock In: NIPS. (2016)

\bibitem{Girdhar2016}
Girdhar, R., Fouhey, D.F., Rodriguez, M., Gupta, A.:
\newblock Learning a predictable and generative vector representation for
  objects.
\newblock In: ECCV. (2016)

\bibitem{Wu2016}
Wu, J., Zhang, C., Xue, T., Freeman, W.T., Tenenbaum, J.B.:
\newblock Learning a probabilistic latent space of object shapes via {3D}
  generative-adversarial modeling.
\newblock In: NIPS. (2016)

\bibitem{Liu2017}
Liu, J., Yu, F., Funkhouser, T.A.:
\newblock Interactive {3D} modeling with a generative adversarial network.
\newblock In: 3DV. (2017)

\bibitem{Gwak2017}
Gwak, J., Choy, C.B., Garg, A., Chandraker, M., Savarese, S.:
\newblock Weakly supervised generative adversarial networks for {3D}
  reconstruction.
\newblock In: 3DV. (2017)

\bibitem{Riegler2017}
Riegler, G., Ulusoy, A.O., Geiger, A.:
\newblock {OctNet}: Learning deep {3D} representations at high resolutions.
\newblock In: CVPR. (2017)

\bibitem{Wang2017}
Wang, P.S., Liu, Y., Guo, Y.X., Sun, C.Y., Tong, X.:
\newblock {O-CNN}: {O}ctree-based convolutional neural networks for {3D} shape
  analysis.
\newblock In: SIGGRAPH. (2017)

\bibitem{Hane2017}
H{\"{a}}ne, C., Tulsiani, S., Malik, J.:
\newblock Hierarchical surface prediction for {3D} object reconstruction.
\newblock In: 3DV. (2017)

\bibitem{Maxim2017}
Tatarchenko, M., Dosovitskiy, A., Brox, T.:
\newblock Octree generating networks: Efficient convolutional architectures for
  high-resolution {3D} outputs.
\newblock In: ICCV. (2017)

\bibitem{Lun3DV2017}
Lun, Z., Gadelha, M., Kalogerakis, E., Maji, S., Wang, R.:
\newblock {3D} shape reconstruction from sketches via multi-view convolutional
  networks.
\newblock In: 3DV. (2017)

\bibitem{Sinha2017}
Sinha, A., Unmesh, A., Huang, Q., Ramani, K.:
\newblock {SurfNet}: {G}enerating {3D} shape surfaces using deep residual
  network.
\newblock In: CVPR. (2017)

\bibitem{Yumer2016}
Yumer, M.E., Mitra, N.J.:
\newblock Learning semantic deformation flows with {3D} convolutional networks.
\newblock In: ECCV. (2016)

\bibitem{Chen2017}
Kong, C., Lin, C.H., Lucey, S.:
\newblock Using locally corresponding {CAD} models for dense 3{D}
  reconstructions from a single image.
\newblock In: CVPR. (2017)

\bibitem{Pontes2017}
Pontes, J.K., Kong, C., Eriksson, A., Fookes, C., Sridharan, S., Lucey, S.:
\newblock Compact model representation for {3D} reconstruction.
\newblock In: 3DV. (2017)

\bibitem{kuryenkov2017deformnet}
Kurenkov, A., Ji, J., Garg, A., Mehta, V., Gwak, J., Choy, C.B., Savarese, S.:
\newblock {DeformNet:} {F}ree-form deformation network for 3d shape
  reconstruction from a single image.
\newblock Volume abs/1708.04672. (2017)

\bibitem{rubner2000earth}
Rubner, Y., Tomasi, C., Guibas, L.J.:
\newblock The earth mover's distance as a metric for image retrieval.
\newblock International journal of computer vision \textbf{40}(2) (2000)
  99--121

\bibitem{howard2017mobilenets}
Howard, A.G., Zhu, M., Chen, B., Kalenichenko, D., Wang, W., Weyand, T.,
  Andreetto, M., Adam, H.:
\newblock Mobilenets: {E}fficient convolutional neural networks for mobile
  vision applications.
\newblock arXiv preprint arXiv:1704.04861 (2017)

\bibitem{ImageNet}
Deng, J., Dong, W., Socher, R., Li, L.J., Li, K., Fei-Fei, L.:
\newblock {I}mage{N}et: {A} large-scale hierarchical image database.
\newblock CVPR (2009)  248--255

\bibitem{ShapeNet}
Chang, A.X., Funkhouser, T., Guibas, L., Hanrahan, P., Huang, Q., Li, Z.,
  Savarese, S., Savva, M., Song, S., Su, H., Xiao, J., Yi, L., Yu, F.:
\newblock {ShapeNet: An Information-Rich 3D Model Repository}.
\newblock Technical Report arXiv:1512.03012 [cs.GR] (2015)

\end{thebibliography}
\end{document}